\documentclass[a4paper, 12pt]{article}

\usepackage{fancyhdr}
\usepackage{amsmath}
\usepackage{graphicx}
\usepackage{url}
\usepackage{amsbsy,amssymb} 
\usepackage{amsmath}
\usepackage{abstract}
\usepackage{hyperref}
\usepackage{color}
\usepackage{authblk}

\usepackage{fancyvrb}

\usepackage[all]{xy}

\usepackage{rotating}

\usepackage[caption = false]{subfig}

\usepackage{lscape}

\newtheorem{definition}{Definition}

\newtheorem{observation}{Observation}

\title{Natural measures of alignment}
\author[1,2]{R. A. Kycia\footnote{E-mail: \texttt{kycia.radoslaw@gmail.com}}}
\author[2]{Z. Tabor}
\affil[1]{Department of Mathematics and Statistics, Masaryk University, Brno, Czechia}
\affil[2]{Cracow University of Technology, Faculty of Physics, Mathematics and Computer Science, PL-31155, Krak\'ow, Poland}

\date{}
\begin{document}
\maketitle
\begin{abstract}
\noindent
Natural coordinate system will be proposed. In this coordinate system alignment procedure of a device and a detector can be easily performed. This approach is generalization of previous specific formulas in the field of calibration and provide top level description of the procedure. A basic example application to linac therapy plan is also provided.
\end{abstract}


\section{ Introduction }

There is many approaches to measurement of discrepancy between ideal case and those obtained from real experiment. Short overview of the subject can be found in \cite{Callibration}.

This paper redefines the measurement of the quality of alignment using some kind of specific natural coordinates system based on induced group action in the detector space. It uses approach from category theory \cite{Conceptual, Spivak, Spivak2, CategoryForWorking} and abstract algebra \cite{AlgebraChapter0}, i.e., the calibrated device is accessible only by the 'morphisms' - the set of gauges that can be changed in order to adjust the configuration. This description does not go into device details. These gauge degrees can be represented as a group action on the device, which change its configuration. In our approach we focus only on continuous groups - Lie groups \cite{Humphreys}, e.g., rotation of device, position change, beam intensity etc. These gauge degrees of freedom projected on the detector give some natural coordinates that can be used to made calibration. The number of 'projected' coordinates depends on detector geometry and position. It is a 'maximal set' that can be used for perfect calibration using available gauges only. This is common task in practice as usually no manipulation of the device apart using the gauges is allowed.

This paper is significant generalization of the approach from \cite{Callibration}. The method presented in \cite{Callibration} is in many cases suboptimal as the number of parameters for optimization is selected in most cases arbitrarily. This leads to overdetermined optimization problem. In this context our method presents one specific case, based on geometry of the problem, maximal number of available degrees of freedom and reduction of degrees of gauges by projection. We start from small number of general assumptions on geometry and basic facts from category theory and representation theory of groups. This leads to somehow unique (up to coordinate selection) method that can be extended adding 'new degrees' of gauges, i.e. group action, and therefore optimization parameters. This approach can be seen as a scheme for generating calibration methods. This abstract description of the calibration process also allows to identify points where the procedure can fail and therefore gives exact and precise suggestions how to avoid these problematic points.

The paper is organized as follows: In the next section general description of natural coordinate system in detector space is presented. Then the idea of alignment measure is given. Finally in the last section some specific example applied to the cancer therapy plan is presented.

\section{ General considerations }
In this section general considerations on the device and detector setup using Lie groups approach \cite{Humphreys} will be presented in two consecutive subsections. The presentation uses continuous groups, however discrete groups \cite{AlgebraChapter0} can be also used in analogous way.

\subsection{ The device }
It is assumed that there is an imaging device or generally a device which output can be registered by some detector. The device has some degrees of freedom and can be moved along some directions using  gages. This movement can be described by the action of some Lie group (continuous group) $\mathcal{G}$ on the space of parameters of the device, see \cite{Humphreys} for review of Lie groups, and \cite{AlgebraChapter0} for group action.

The device can be seen as a set of parameters in some space that uniquely gives configuration at a given instant of time. These parameters can have either geometrical meaning like position or angle, or intrinsic meaning like the intensity of the imaging beam. Therefore, we have the first observation
\begin{observation}
There is iso mapping from the device to the parameter space
\begin{center}
\xymatrix{ D \ar[r]^{iso} & P, } 
\end{center}
where $D$ is the device and $P$ corresponding parameter space. 
\end{observation}
Usually parameter space is a simple Cartesian product, e.g., for position of a point of the detector one can use $\mathbb{R}^{3}$ and for an angle variable one can choose the circle $S^{1}$, and therefore the full parameter space is $\mathbb{R}^{3} \times S^{1}$. The dimension of the space of parameters depends on the assumed level of accuracy of description. In a finer model the iso maps the device to the bigger dimensional space of parameters. The map is called 'iso' as it is an isomorphism under assumed level of description accuracy.

The gauges allow to change configuration of the device and they have properties of group - there is identity (no change in configuration) and composition law of two gauges change holds (applying two gauges we get the result which is a third gauge). For inverse of action of gauge it usually exists as we can return to the starting setup of the device. Sometimes the composition law exists only for small gauge changes however for large changes may or may not exists due to some constraints in movements or safety. Therefore all gauge changes usually have group property which is sometimes limited. We will restrict ourselves to the cases where the full group structure exists. The constraints can be derived from our considerations by removing some elements of a group.
\begin{observation}
Change in device configuration can be seen as a group action in the device parameter space as
\begin{equation}
 g.p=p',
\end{equation}
where $g\in \mathcal{G}$ is the group element corresponding to the gauge action and $p,p'\in P$ are configuration before and after transformation respectively.
\end{observation}

More generally, for continuous or sequential but discrete change in the configuration of the device we have the following
\begin{observation}
 If there is a continuous curve in the group $c(t): \mathbb{R} \rightarrow \mathcal{G}$ parametrized by e.g. time or some parameter from ordered set, then we can model a change in device parameter space as a curve in its parameter space. The change in configuration in time can be defined by
\begin{equation}
 p(t):=c(t).p, \quad p\in P,
\end{equation}
where $p(t)$ is a curve in the device parameter space.
\end{observation}
In this approach for discrete groups the 'curve' is a sequence of element action $c=g_{t_1}g_{t_2}\ldots g_{t_n}$.

The next subsection describes model of the detector.

\subsection{ The detector }
The detector can be seen as a piece of some space, e.g., rectangle cut from a plane or a finite cylinder homeomorphic to $S^{1}\times [0;1]$ cut from infinite cylinder $S^{1}\times \mathbb{R}$.

The device when set at some values of parameters in the device configuration space, gives an image $I$ in the detector, which can be seen, e.g., as an array of pixel values.

The next observation connects the group of the device with induced group structure in the detector, namely,
\begin{observation}
The action of the group $\mathcal{G}$ on the device $D$, induces surjectively the action of the group on the image $I$ in the detector. This action can be seen by the detector space as some image  $S(\mathcal{G})$ of $\mathcal{G}$. The map is usually non-bijective. The kernel of $S$ map describes 'the loss in coordinates' during the projection on the detector space. Therefore $|ker(S)|$ - the rank of $ker(S)$, shows how many degrees of freedom is removed by passing from $\mathcal{G}$ to $S(\mathcal{G})$. Therefore it is vital to select the detector that minimise $|ker(S)|$. 
\end{observation}
It is crucial observation and also the most important assumption in the paper - the group of the device induces a structure in the detector which is also a group. It is reasonable assumption and without it the remaining part of the paper is not valid.

In addition, we assume that there is no other elements that acts on the image $I$ in the detector than those induced by $S(\mathcal{G})$. In other words, we can manipulate of the image $I$ only by gauges. 

Therefore, we have following
\begin{observation}
 The group $\mathcal{G}$ induces in the detector the action of the group $S(\mathcal{G})$ on the image, i.e,
 \begin{equation}
  I'=s.I, \quad s \in S(\mathcal{G}),
 \end{equation}
where $I'$ is the transformed image of $I$.
\end{observation}

Summing up, we have the following observation
\begin{observation}
 The relation between device and detector induced groups can be visualized in the following sequence
\begin{center}
\xymatrix{ \mathcal{G} \ar[r]^S & S(\mathcal{G}).}
\end{center}
It can be made more explicit by passing to the quotient space. According the first isomorphism theorem for groups \cite{AlgebraChapter0} the $S$ splitting has he form
\begin{center}
\xymatrix{ \mathcal{G} \ar@/_2pc/[rr]_{S} \ar@{->>}[r]^{q_{S}} & \mathcal{G}/ker(S)  \ar[r]^{\tilde{S}} & im(S), }
\end{center}
where $q_{S}$ is the quotient map and $\tilde{S}$ is defined uniquely by $S=\tilde{S}\circ q_{S}$. Here $\tilde{S}$ is an isomorphism.
\end{observation}

As every kernel is a normal subgroup \cite{AlgebraChapter0}, we can therefore estimate how many different types of detectors can be constructed, i.e., we have

\begin{observation}
 The number of different kind of detectors that can be constructed according to the device gauge group $\mathcal{G}$ is given by the multiciplity of normal subgroups in $\mathcal{G}$. The dimension of the selected normal group describes the degrees of freedom that are not visible in the detector when acting on the device by $\mathcal{G}$ and this defines 'a type of the detector'.
\end{observation}

If the group is a Lie group then it forms a manifold and locally we can introduce coordinates. Then using previous observations one can introduce coordinates in the detector space. Taking into account that the quotient of the Lie group by normal subgroup is also a Lie group \cite{Humphreys} we arrive at
\begin{definition}
 \textbf{Natural coordinates in a detector space} are coordinates on the manifold that is a subgroup $\mathcal{G}/ ker(S)$.
\end{definition}
In addition, if the group is a direct product of Lie group and some discrete group then in the discrete part we can also select some (discrete) sets of values for coordinates.

These natural coordinates can be used to measure discrepancy in alignment/calibration in a natural way. This will be the subject of the next section.

\section{ Alignment measurement}

In order to measure alignment in the detector space the appropriate metric has to be defined. From mathematical point of view, if we interpret the image in the detector as a set $X$ (which can be considered as a vector space), then the metric is a function $|| \cdot ||:X \rightarrow \mathbb{R}_{+}$ that fulfils the following well-known properties for $x,y \in X$:
\begin{itemize}
 \item {$||x||=0 \Longleftrightarrow x=0$; }
 \item {$||ax||=|a| ||x||$ for $a \in \mathbb{R}$;}
 \item {$||x+y|| \leq ||x||+||y||$;}
\end{itemize}
However, for some purposes some of the above conditions can be relaxed.

As we will see in a while, the most important concept in selecting proper metric is the sphere of the metric, i.e.,
\begin{definition}
 Spheres (isometric surfaces) for a metric $|| \cdot ||$ are defined as
 \begin{equation}
  C(c)=\{ x\in X | ||x|| = c \},
 \end{equation}
 for the radius $c \in \mathbb{R}_{+}$. Each choice of $c$ gives a sphere for a given metric.
\end{definition}

Since, the image in the detector is transformed according to the induced group $S(\mathcal{G})$, therefore, the important issue is to maximize this action to be not orthogonal to the spheres of selected metric. We have

\begin{observation}
 Metric for calibration measurement should be selected in such a way that the spheres of the metric has the least as possible symmetries given by the induced group $S(\mathcal{G})$.
 
 The worst case scenario is to select a metric which spheres have a whole group $S(\mathcal{G})$ as its stabilizer, i.e., the subgroup $G_{x}=\{g \in S(\mathcal{G}) | g.x=x\}$.
\end{observation}

Intuitively, when group moves a sphere of the metric to itself then we cannot see difference before and after its action, and as a result, calibration cannot be performed.

For some specific cases, e.g., when the metric is given by some smooth function(s) or algebraic ones then there is a notion of a tangent space to the metric spheres $C(c)$ for a fixed $c$. In this case the optimally selected metric is such that the tangent spaces to the spheres do not contain vectors defining $S(\mathcal{G})$ action. In this smooth case one can define how good the selected metric is by noting how many dimensional linear subspace spanned by the vectors given by the group action $S(\mathcal{G})$ lies in the tangent spaces of the spheres $C(c)$ of the metric. This dimension can depend on $c$.

Lets focus for a moment into details of detector construction. Usually detectors are constructed as a set of simple elements (pixel detectors) organized in some geometric structure. Assume therefore that the detector is a set of pixels and the image is a matrix of pixel measurements $I=[I_{i_{1}\ldots i_{n}}]$, which depending on the arrangement of pixels. It can be a scalar (single pixel), a vector (line of pixels) or a tensor in a case of an array or volume pixels of some arrangement.

Now in the space of pixel detectors one can introduce some usual norm. As it was summarized in \cite{Callibration} for defining of such measures one can select comparison on the level of pixels (pixel-level techniques) or features of an image (feature-based techniques). It is arbitrary choice in every algorithm and depends on the characteristics that have to be calibrated/aligned.

For purpose of presentation the simplest pixel-based approach is selected, i.e., one norm from the standard family of them
\begin{equation}
 ||I||_{k}:=\left( \sum _{i_{1},\ldots, i_{n}} |I_{i_{1}\ldots i_{n}} |^{k} \right)^{1/k},
 \label{Eq.k-metric}
\end{equation}
which induces the distance function between two images $I^{1}$ and $I^{2}$
\begin{equation}
 d_{k}(I^{1},I^{2}) = ||I^{1}-I^{2}||_{k}.
\end{equation}

We assume that the pixel granularity can be neglected and therefore we do not restrict $S(\mathcal{G})$ to discrete group. The limit of continuous detector can be obtained. For instance it can be some interpolation procedure between pixels as some convergence process that starts from big size (aggregates) of pixels and iteratively change the size ad infinitum. This is not too restrictive assumption - all results can be 'discretized' if desired.

In the next step the definition of some 'measure of calibration' should be provided. In order to measure 'fitness' requires the existence of a standard measurement $I$ that we expect to have when the device is ideally calibrated, and the real measurement $T$. We want to measure the distance between $T$ and $I$. Let us define function $F:S(\mathcal{G}) \rightarrow R_{+}$ by the formula
 \begin{equation}
  F(s)= || I - s.T||.
  \label{Eq.OptimizingFunctionGeneral}
 \end{equation}
It measures the discrepancy of the ideal image $I$ from the real image $T$ altered by the group element $s \in S(\mathcal{G})$ in the detector measurement. This motivates 
\begin{definition}
 The \textbf{perfect alignment} for selected metric (e.g., using (\ref{Eq.k-metric}) with some fixed value $k$), is the $\bar{s} \in S(\mathcal{G})$ such that $F(\bar{s})=min$. This corresponds to the element $\bar{g}_{S} \in \mathcal{G}/ker(S)$ and therefore to the whole coset of $\mathcal{G}$ of the form $\bar{g} ker(S)$ for some $\bar{g} \in \mathcal{G}$.
\end{definition}
Therefore, the perfect calibration can be made by the device gauge up to the $ker(S)$ subgroup that describes the design of the detector.

In order for the perfect calibration to exists some assumption on the compactness of the group $S(\mathcal{G})$ should be imposed. However they are natural in many cases as gauges are some compact topological space like circles. Sometimes the gauges have some finite set of values, which is also a compact (group) in usual topology. If there is no compactness, then the minima do not have to be attained by any set-up of gauges but e.g. approximated by the limiting set-up of gauges.

In addition, even under the compactness assumption more than one perfect calibration exists, e.g., when the kernel of $S$ mapping is non-empty then the whole coset is a perfect calibration.

Up to now no specific assumptions on the groups under considerations were made and these considerations are the most general. For obtaining specific formulas for (\ref{Eq.OptimizingFunctionGeneral}) in the case of continuous groups the local coordinates have to be introduced. For the map $\mathcal{G} \xrightarrow{S} S(\mathcal{G})$ assume that $dim(S(\mathcal{G})) = N$, and moreover that $dim(\mathcal{G})=M$. Therefore on the group manifold $\mathcal{G}$ we can locally introduce the coordinates $\{y_{1}, \ldots, y_{M}\}$. One can introduce them in such a way that the last $M-N = dim(ker(f))$ belongs to $ker(S)$. Then in the detector space, $S(\mathcal{G})$ can be parametrized locally by $\{y_{1},\ldots, y_{N}\}$. Therefore an element $s \in S(\mathcal{G})$ depends functionally on the coordinates $s(y_{1},\ldots, y_{N})$. These coordinates will be used in the further considerations.

We use coordinate chart for the Lie group $S(\mathcal{G})$ seen as a manifold and a norm $|| \cdot ||$ in the detector space.

\begin{definition}
 \textbf{Natural coordinates of image $T$ with respect to the standard image $I$ in the detector} is a set of values $\{\tilde{y_{1}},\ldots,\tilde{y_{N}}\}$ that minimize the function
 \begin{equation}
  F(\tilde{y_{1}},\ldots,\tilde{y_{N}}) = min,
 \end{equation}
 where 
 \begin{equation}
  F(y_{1},\ldots,y_{N})= || I - s(y_{1},\ldots, y_{N}).T||.
  \label{Eq.OptimizingFunctionCoordinates}
 \end{equation}
\end{definition}
The coordinates of $T$ correspond to the group action that gives minimal deviation of $T$ from $I$. These are not coordinates in the strict sense as there can be more than one set of their values corresponding to the minimum of $F$. In the case when there standard image is invariant in the detector under some subgroup action of $S(\mathcal{G})$ then the set of minima can be set of coordinates parametrizing of this subgroup/submanifold restricted to the detector shape. In such a case we have to introduce some metric that measure the distance of this set from a selected point that can be placed at the origin and characterize optimal configuration of the device. 

The formula (\ref{Eq.OptimizingFunctionCoordinates}) is proposed in \cite{Callibration} for the optimization problem of finding calibration. In our paper it results from very general and natural assumptions on the nature of calibration process.

The last step is to define the distance of $T$ to optimal alignment $I$.
\begin{definition}
 Associate with group coordinates $y=\{y_{1},\ldots,y_{N}\}$ the set of nonnegative weight $w=\{w_{1},\ldots,w_{N}\}$ that can be always normalized $\sum_{l=1}^{N}w_{i} = 1$. \textbf{Natural alignment distance} can be defined as the functional
 \begin{equation}
  d_{nat}(I,T) = \sum_{i=1}^{N} w_{i} |y_{i}|^{2},
 \end{equation}
 or more generally
 \begin{equation}
  d_{nat}(I,T) = || \{w_{1}y_{1},\ldots,w_{N}y_{N}\} ||.
 \end{equation}
 for some metric $|| \cdot ||$, which, in general, is different from the metric used to define (\ref{Eq.OptimizingFunctionCoordinates}).
\end{definition}
These weights measure importance of a given group coordinates in the alignment process and when no additional information is given, then they can be uniformly distributed, i.e., $w_{i}=\frac{1}{N}$.

In case that the natural coordinates of $T$ image from $I$ give a set then the natural alignment distance is the distance of the set from the origin that have some special meaning for the device setup.

In the next section we present the abovementioned procedure of construction calibration procedure that is applied to measurement of therapy plan in radiotherapy. We want to emphasize that the method is not limited only to one class of examples and can be applied to any other calibration process.

\section{ Example - therapy plan for linac devices }

For medical linac device before medical treatment is performed, there test that if the sweep of therapeutic beam around cylindrical pixel detector agrees with therapeutic plan has to be made. We will construct quality measure for this test using results from the previous sections.

The device consist the gantry - source of ionising radiation, that can rotate around the patient table. In the calibration the detector has a shape of cylinder that is covered by pixels registering radiation dose.

In order to identify the group action we assume that the gantry can rotate around the axis of the cylindrical detector, move along the cylinder axis, go further and closer and change intensity. It is simplified presentation only for better understanding of the algorithm. In general, additional degrees of freedom that represent misalignment of gantry according to the cylinder axis have to be introduced.

\begin{figure}
\centering
 \includegraphics[width = 0.7\textwidth]{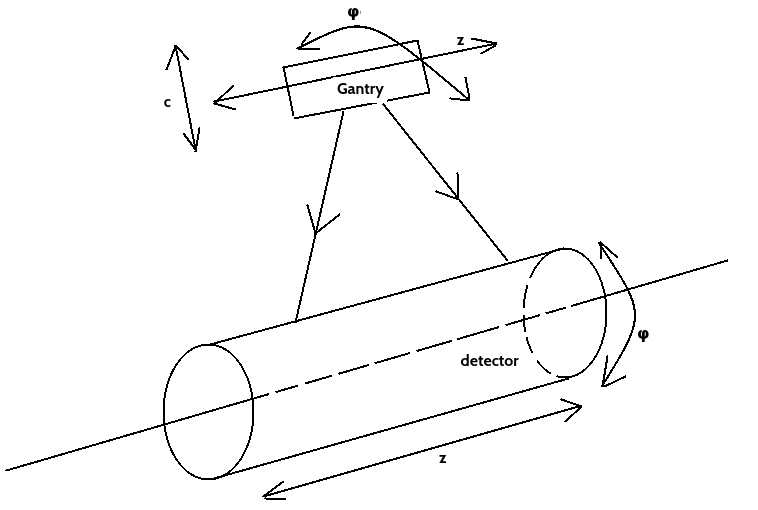}
 \caption{Gantry-detector degrees of freedom. The movement of gantry along $c$ corresponds to the scaling of the image in the detector. Internal intensity $i$ parameter is not depicted as it is not geometric degree of freedom/group action.}
 \label{Fig:TherapyPlan}
\end{figure}

The coordinates in the system are visualized in Fig. \ref{Fig:TherapyPlan}.
We can distinguish in the detector space the following group induced coordinates
\begin{itemize}
 \item {\textbf{geometric} - cylindrical coordinates on the surface of the cylinder $\phi$, $z$;}
 \item {\textbf{intrinsic that results from geometric projection of degrees of freedom of the device} - scaling $c$ of the detector image projected to the cylindrical shape of the detector that is associated in radial movement of the gantry;}
 \item {\textbf{intrinsic} - intensity $i$ of the beam acting multiplicatively on the given intensity measured in a pixel, i.e., $[s(i).T]_{ij}= i\times T_{i,j}$;}
\end{itemize}
For example $s(\phi,z=0, c=0, i=1).T$ gives rotated by $\phi$ the image $T$ around the cylinder detector and $d(\phi=0,z=0, c=0,i).T$ is the image with $i$ times higher intensity detected in every pixel.

Summing up, the group of the device $\mathcal{G}$ contains a product of independent abelian subgroups for every degree of freedom. When the detector contains a single pixel then the scaling and intensity do not commute (see below). Each abelian subgroup is normal and therefore classes of the detectors contain devices that can measure all degrees of freedom or has some of them omitted. 

For example, constructing the cylindrical detector and placing it according to the axis of rotation of the device the induced group of the image $S(\mathcal{G})$ has trivial kernel. However, when detector is a line segment of pixels and the $ker(S)$ is selected to be a rotation subgroup then placing it along the main axis (perpendicular to the rotation $\phi$ coordinates of the normal group) we can measure $z$ position of the device and intensity, as well as scaling. If such line detector is not aligned according to the main axis then it measures one dimensional submanifold in the $\mathcal{G}$ that is described by the equation $\phi(z)$, and therefore $ker(S)$ is still nontrivial and and has rank one. However, coordinates are not chosen optimally - adjusted to some normal subgroup. Trivially, if the detector is a single pixel then it can measure intensity only at given point and the normal subgroup of the kernel contains everything apart of intensity gauge and scaling which are indistinguishable by this detector.

Assume that we selected cylindrical detector. Now we calculate the coordinates $\{\tilde{\phi}, \tilde{z}, \tilde{c}, \tilde{i}\}$ by minimalizing the function
\begin{equation}
 F_{k}(\phi,z,c,i)=||I - s(\phi,z,c,i).T||_{k},
\end{equation}
where $I$ is a standard therapy plan crafted for the patient, and $T$ is an image which was obtained in the detector by realizing the plan by the device.

Note, e.g., that the detector has natural translation symmetry in $\phi$ coordinate direction, and therefore, if $I$ poses also this symmetry then $F_{k}$ has the line $\phi$ as location of minima. One can then choose arbitrary, e.g., the smallest value, i.e., $\phi=0$ as a coordinate. This is the ambiguity of the natural coordinates that were mentioned above. 

Finally, the priority of group coordinates can be set be selecting the set of weights for coordinates $\{w_{\phi},w_{z},w_{c},w_{i}\}$ which sum is normalized to unity. For example, the highest weight can be associated with intensity $w_{i}$ which is connected with dose taken during the radiotherapy. The number
\begin{equation}
 d_{k,nat}(I,T) = w_{\phi}\tilde{\phi}+w_{z}\tilde{z}+w_{c}\tilde{c}+w_{i}\tilde{i},
\end{equation}
describe how far is the obtained plan $T$ from the crafted plan $I$ for a patient with specific set of weights in natural coordinates.

\section{ Summary }
General method for checking quality of calibration/alignment for the system imaging device using natural coordinate was outlined. It bases heavily on the allowed movements of the device given by a Lie group and its induced group in the detector device. The method allows to predict what types of detectors are possible based on the normal subgroups and how common mistakes can be avoided. Illustrative example for radiological therapy plan was presented. 

The method is designed to be general and applicable to many device-detector systems and therefore allows to control quality of various devices.

\section*{ Acknowledgements }
RK thanks Valentin Lychagin and Josef Silhan for inspiring discussions on Lie groups and differential geometry. This research was supported by the grant POIR.04.01.04-00-0014/16 of The National Centre of Research and  Development. RK was supported by GACR Grant 17-19437S, and the grant MUNI/A/1138/2017 of Masaryk University.




\end{document}